\documentclass[conference]{IEEEtran}
\usepackage{amsfonts}
\usepackage{graphicx}
\usepackage{algorithm}
\usepackage{algorithmic}
\usepackage{amssymb}
\usepackage{amsmath}
\usepackage{setspace}
\usepackage{xcolor}
\usepackage{graphicx}
\usepackage{subfigure}
\usepackage{lipsum}
\usepackage{mathtools}
\usepackage{textcomp}

\begin{document}
\title{Learning-Based Resource Management in Integrated Sensing and Communication Systems}

\author{\IEEEauthorblockN{Ziyang Lu,  M. Cenk Gursoy,  Chilukuri K. Mohan,  Pramod K. Varshney} 
%
\IEEEauthorblockA{Department of Electrical Engineering and Computer Science, Syracuse University, Syracuse NY, 13066 
\\
%
\{zlu112, mcgursoy, ckmohan, varshney\}@syr.edu
}}


\maketitle

\begin{abstract}\let\thefootnote\relax\footnote{This work was supported in part by the National Science Foundation under Grant CNS-2221875.}
In this paper, we tackle the task of adaptive time allocation in integrated sensing and communication systems equipped with radar and communication units. The dual-functional radar-communication system's task involves allocating dwell times for tracking multiple targets and utilizing the remaining time for data transmission towards estimated target locations. We introduce a novel constrained deep reinforcement learning (CDRL) approach, designed to optimize resource allocation between tracking and communication under time budget constraints, thereby enhancing target communication quality. Our numerical results demonstrate the efficiency of our proposed CDRL framework, confirming its ability to maximize communication quality in highly dynamic environments while adhering to time constraints.
\end{abstract}

\section{Introduction}

\subsection{Background}
\subsubsection{Cognitive Radar}
Radar technology, integral to various applications in environmental sensing, space exploration, navigation, and traffic control, has become increasingly important with the emergence of autonomous vehicles and drones. Efficient allocation of radar resources in challenging environments has thus been a key focus of research. Traditional optimization methods for resource allocation are discussed in \cite{orman1996scheduling} and \cite{butler1997resource}, offering foundational approaches to the problem.

The literature also explores game-theoretical methods for radar resource management. Authors in \cite{deligiannis2017game} address power allocation in multi-radar systems, treating it as a non-cooperative game and analyzing the Nash equilibrium and its convergence. This approach highlights the strategic interactions in radar systems and their implications for resource management.

Cognitive radar, a field at the intersection of radar technology and AI, has been rapidly evolving. In particular, cognitive radar leverages machine learning, game theory, and cognitive radio techniques to enhance radar performance in dynamic and nonstationary environments. This aspect of radar technology is covered in depth in \cite{charlish2020development} and \cite{haykin2006cognitive}, illustrating how cognitive radar systems use AI to adapt and make informed decisions. These systems are designed to continuously update their understanding of the environment, enabling more effective responses to changing conditions.

The authors in \cite{charlish2020development} provide a comprehensive overview of cognitive radar systems, focusing on aspects like signal processing, dynamic feedback, and information preservation. This area is pivotal in developing multifunctional radar systems capable of tasks like multi-target tracking and electronic beam steering. The literature underscores the transformative impact of AI and cognitive techniques on radar technology, paving the way for more adaptive and intelligent systems capable of operating in increasingly complex scenarios.

\subsubsection{Integrated Sensing and Communication (ISAC)}
ISAC represents a paradigm shift in wireless networks, integrating sensing and communication functionalities within a single platform. This integration is crucial in the era of 5G and beyond, offering enhanced efficiency and capabilities in various applications. The convergence of sensing and communication technologies, particularly with the advent of millimeter-wave and massive MIMO technologies, has led to significant advancements in ISAC research and applications as illustrated in \cite{liu2022survey}. 

ISAC is also anticipated to play a pivotal role in the evolution of 6G wireless networks. The dense cell infrastructures in future networks provide a unique opportunity to construct perceptive networks that integrate sensing directly into the communication process. The incorporation of ISAC in future cellular systems, WLAN, and V2X networks is expected to bring substantial integration and coordination gains \cite{liu2022integrated}.

Cognitive radar is a system that adapts its operating parameters based on the environment and aligns closely with the principles of ISAC. It involves intelligent decision-making and adaptive sensing, which are key components of ISAC systems. In the context of ISAC, cognitive radar can benefit from the shared use of communication and sensing resources, leading to more efficient and responsive systems, hence motivating the work in this paper.

\subsection{Related Work}
A number of recent research efforts have focused on resource management in radar systems. For example, the study in \cite{schope2021constrained} tackles the challenge of time allocation in tracking multiple targets using the extended Kalman filter approach, within the context of partially observable Markov decision processes and policy rollout techniques. Another line of research in \cite{de2021radar} utilizes model predictive control (MPC) for allocating time in radar systems. The simulation findings reveal that both policy rollout and MPC are effective in optimizing the revisit interval and dwell time, leading to reduced estimation variance. These methods, being offline, base their decisions on pre-existing knowledge, such as statistical data about measurement and maneuverability noise. However, given the nonstationary nature of radar environments, these offline approaches might struggle with rapid environmental changes.

To address such dynamic scenarios, online strategies capable of adapting to environmental shifts are preferable. Deep reinforcement learning (DRL) has gained prominence in training agents for complex decision-making tasks, exhibiting impressive results in \cite{mnih2015human}. DRL's model-free nature and its inherent features make it apt for dynamic radar application challenges. In fact, DRL has been increasingly incorporated into radar-related decision-making. For instance, authors in \cite{thornton2020deep} employed DRL for spectrum allocation in multi-target detection, demonstrating improved detection performance and reduced interference with other wireless systems. Additional DRL applications in radar contexts have been explored in \cite{selvi2020reinforcement} and \cite{meng2021deep}.

\subsection{Contributions}
This work addresses the time allocation problem in ISAC systems, aiming to optimize the balance between tracking and communication for enhanced overall communication quality with the targets. Our main contributions are the following:

\begin{itemize}
    \item \emph{Formulation of the time allocation problem as a constrained optimization problem}: This involves developing a strategy for the ISAC system to balance tracking and communication phases, thereby maximizing target communication quality.

    \item \emph{Design of a constrained deep reinforcement learning (CDRL) framework to solve the optimization problem}: We detail the method for simultaneously learning the parameters of the deep Q-network (DQN) and the dual variable.

    \item \emph{Numerical analysis demonstrating the CDRL framework's effectiveness}: The results validate that our approach successfully learns an efficient time allocation strategy while adhering to the constraints of the available time budget.
\end{itemize}

\section{Radar Tracking Model}

\subsection{Target Motion Model}

At time slot $t$, the state of a target is captured as $\mathbf{x_t} = [x_t, y_t, \dot{x}_t, \dot{y}_t]^T$, where $(x_t, y_t)$ indicates the target's present position, and $(\dot{x}_t$, $\dot{y}_t)$ represents its horizontal and vertical velocities. Under a constant velocity model during the revisit period, the target's state transitions from $\mathbf{x_t}$ to
\begin{equation}
\mathbf{x_{t+1}} = \mathbf{F_t} \mathbf{x_{t}} + \mathbf{w_t},
\end{equation}
with $\mathbf{F_t} \in \mathbb{R}^{4\times4}$ being the state transition matrix, expressed as
\begin{equation}
	\mathbf{F_t} = \begin{bmatrix}
		1&  0&  T_t&  0\\
		0&  1&  0&  T_t\\
		0&  0&  1&  0\\
		0&  0&  0&1  \\
	\end{bmatrix}
\end{equation}
where $T_t$ is the radar system's time interval for tracking at time $t$. Here, $\mathbf{w_t}$ signifies the maneuverability noise, modeled as a multivariate zero-mean Gaussian noise with covariance matrix
\begin{equation}
	\mathbf{Q_t} = \begin{bmatrix}
		T_t^4/4&  0&  T_t^3/2&  0\\
		0&  T_t^4/4&  0&  T_t^3/2\\
		T_t^3/2&  0&  T_t^2&  0\\
		0&  T_t^3/2&  0&T_t^2  \\
	\end{bmatrix} \sigma_{w}^2
\end{equation}
where $\sigma_{w}^2$ represents the variance of the maneuverability noise at time $t$.

\subsection{Measurement Model}

The radar system acquires measurements of the range $r$ and azimuth angle $\theta$ of the target for location estimation. The measurement vector, denoted by $\mathbf{z_t}$, and the non-linear mapping function from $\mathbf{x_t}$ to $\mathbf{z_t}$, denoted by $h(\cdot)$, establish the relation between the state and measurement vectors as follows:
\begin{equation}
	\mathbf{z_t} = h(\mathbf{x_t}) + \mathbf{v_t} = \left[\sqrt{x_t^2+y_t^2}, \quad \text{tan}^{-1}\left(\frac{y_t}{x_t}\right)\right]^T + \mathbf{v_t},
\end{equation}
where $\mathbf{v_t}$, the measurement noise vector at time $t$, comprises the range measurement noise $v_{r,t}$ and angle measurement noise $v_{\theta,t}$, both modeled as zero-mean Gaussian noise with variances $\sigma_{r,t}^2$ and $\sigma_{\theta,t}^2$. The radar's assumed location is at the Cartesian system's origin.

The measurement noise variance dynamically correlates with the signal-to-noise ratio $\text{SNR}_t$ of the target's reflected radar signal at time $t$. $\text{SNR}_t$ is influenced by the radar's dwell time $\tau_t$, target-radar distance $r_t$, as formulated in \cite{schope2021constrained}, \cite{koch1999adaptive}, and beam misalignment loss $L_{bm}^T$:

\begin{equation}
	\text{SNR}_t(\tau_t, r_t, \Theta, \hat{\Theta}) = \text{SNR}_0\left(\frac{\tau_t}{\tau_0}\right)\left(\frac{r_t}{r_0}\right)^{-4} L_{bm}^T
    \label{SNR}
\end{equation}
where $\text{SNR}_0$, $\tau_0$ and $r_0$ are the reference values of the SNR, dwell time and the distance from target to radar. $L_{bm}^T$ denotes the power loss due to the misalignment of the radar beam direction and the ground truth values of the azimuth angle of the tracked target. In this work, we model the antenna pattern of the radar beam with a cosine response in azimuth. For instance, if we denote the estimated azimuth angle of the target as $\hat{\Theta}$ and the ground truth azimuth angle of the target as $\Theta$. Then the tracking beam misalignment loss $L_{bm}^T$ can be expressed as

\begin{equation}
L_{bm}^T = 
\begin{cases} 
  \text{cos}^j(|\Theta - \hat{\Theta}|) & \text{if } |\Theta - \hat{\Theta}|\leq\frac{\pi}{2}, \\
  0 & \text{if } |\Theta - \hat{\Theta}|>\frac{\pi}{2}
\end{cases}
\label{BM}
\end{equation}
where the antenna pattern of the radar beam is determined by $j$. A larger $j$ corresponds to a narrower beam pattern.

Then, the relationship between the variance of measurement noise and the SNR can be determined as \cite{meikle2008modern}
\begin{equation}
	\sigma_{\bullet,t}^2 = \frac{\sigma_{\bullet,0}^2}{\text{SNR}_t(\tau_t, r_t, \Theta, \hat{\Theta})}
	\label{sigma}
\end{equation}
where $\bullet \in (r, \theta)$. $\sigma_{\bullet,0}^2$ denotes the reference value of the corresponding measurement noise variance. It is worth noting that the variance of the measurement noise decreases when a longer dwell time is allocated to the target or when the target moves closer to the radar according to equations (\ref{SNR}) and (\ref{sigma}).

Note also that the mapping function $h(\cdot)$ between measurements and states is non-linear and hence extended Kalman filter (EKF) is employed in this work. When using EKF, an observation matrix $\mathbf{H_t} \in \mathbb{R}^{2\times4}$ is introduced to linearize the relationship between $\mathbf{z_t}$ and $\mathbf{x_t}$. $\mathbf{H_t}$ is defined as the Jacobian of the measurement function $h(\cdot)$:
\begin{equation}
	\mathbf{H_t} = \frac{\partial h(\cdot)}{\partial \mathbf{x}}\vert_{\mathbf{x_t}} =
	\begin{bmatrix}
		
		\frac{x_t}{\sqrt{x_t^2+y_t^2}} &  \frac{y_t}{\sqrt{x_t^2+y_t^2}} &  0&  0\\
		\frac{-y_t}{x_t^2+y_t^2}&  \frac{x_t}{x_t^2+y_t^2}&  0&  0\\

	\end{bmatrix}.
\end{equation}

Considering independent measurements, the covariance matrix of measurements is given by
\begin{equation}
	\mathbf{R_t} =
	\begin{bmatrix}
		\sigma_{r,t}^2  & 0\\
		0               & \sigma_{\theta, t}
	\end{bmatrix}.
\end{equation}

\subsection{Extended Kalman Filter}
\label{subsec:Kalman}
Kalman filter is a well-known recursive algorithm for estimating the state of a process \cite{welch1995introduction}, which finds its extension in the Extended Kalman Filter (EKF). The EKF, adept for non-linear measurement scenarios like in radar target tracking, operates through two principal phases—prediction and update.

\subsubsection{Prediction Phase}
In this phase, the EKF predicts the target's future state using:
\begin{equation}
	\mathbf{\hat{x}_{t|t-1}} = \mathbf{F_t} \mathbf{x_{t-1|t-1}},
	\label{Kalman1}
\end{equation}
\begin{equation}
	\mathbf{\hat{P}_{t|t-1}} = \mathbf{F_t} \mathbf{P_{t-1|t-1}} \mathbf{F_t}^T + \mathbf{Q_t},
	\label{Kalman2}
\end{equation}
where $\mathbf{\hat{x}_{t|t-1}}$ and $\mathbf{\hat{P}_{t|t-1}}$ are the predicted state and its covariance.

\subsubsection{Update Phase}
Here, the EKF refines its prediction based on new measurements:
\begin{equation}
	\mathbf{K_{t}} = \mathbf{\hat{P}_{t|t-1}}\mathbf{H_t}^T(\mathbf{H_t} \mathbf{\hat{P}_{t|t-1}} \mathbf{H_t}^T + \mathbf{R_t})^{-1},
	\label{Kalman3}
\end{equation}
\begin{equation}
	\mathbf{x_{t|t}} = \mathbf{\hat{x}_{t|t-1}} + \mathbf{K_{t}}(\mathbf{z_{t}} - h(\mathbf{\hat{x}_{t|t-1}})),
	\label{Kalman4}
\end{equation}
\begin{equation}
	\mathbf{P_{t|t}} = (\mathbf{I} - \mathbf{K_t}\mathbf{H_t}) \mathbf{\hat{P}_{t|t-1}},
	\label{Kalman5}
\end{equation}
culminating in the updated state $\mathbf{x_{t|t}}$ and covariance $\mathbf{P_{t|t}}$.

The EKF initializes $\mathbf{x_{t|t}}$ and $\mathbf{P_{t|t}}$ as a zero vector and identity matrix, respectively, iterating to minimize the trace of $\mathbf{P_{t|t}}$.

This enhanced description provides a clearer understanding of the EKF's mechanism, particularly for radar target tracking applications.

\section{Problem Formulation}

\subsection{Constrained Markov Decision Processes (CMDP)}
A constrained Markov decision process (CMDP), as defined by \cite{altman1999constrained}, is characterized by the tuple $(S, A, C, \Theta, T, \mu, \gamma)$. It consists of states $S$, actions $A$, a cost function $C: S\times A \times S \rightarrow \mathbb{R}$, a budget function $\Theta: S\times A \times S \rightarrow \mathbb{R}$, and a transition probability $T: S \times A \times S \rightarrow [0,1]$. The initial state distribution is denoted by $\mu$, and $\gamma \in [0,1)$ represents the discount factor for future rewards and costs. The goal is to devise an optimal policy $\pi:S \times A \rightarrow [0,1]$ that minimizes the discounted sum of future costs (or maximizes the discounted sum reward) while adhering to budget constraints. Mathematically, this is represented as
\begin{equation}
	\begin{aligned}
		\min_{\pi} \quad &  \sum_{m=0}^{\infty}\gamma^m c_{t+m}\\
		\textrm{s.t.} \quad & \sum_{m=0}^{\infty}\gamma^m (\Theta_{t+m} - \Theta_{max})\leq 0.
	\end{aligned}
\end{equation}

\subsection{Time Allocation in Integrated Sensing and Communication}
ISAC systems synergize sensing functionalities, such as radar tracking, with communication capabilities within a unified framework. This study focuses on the time allocation problem in ISAC systems, aimed at optimizing a combined metric of sensing and communication efficiency. This involves dividing the total radar revisit time, denoted by $T_0$, into dwell times $\{\tau_t^n\}_{n=1}^N$ for target tracking, and a communication time $\tau_t^c = T_0 - \sum_{n=1}^N{\tau_t^n}$ for target communication.

The ISAC operation is categorized into two phases: target tracking and communication. In the tracking phase, the radar, using prior target location estimates, emits beams towards the estimated azimuth angles of the targets, capturing echo signals with embedded noisy measurements. Each target $n$ is allocated a specific dwell time $\tau_t^n$. The radar then employs an extended Kalman filter to predict the targets' subsequent positions.

In the communication phase, the radar system first determines the orientation of its communication beams, building upon the previous frame's target location estimations. Accounting for beam misalignment loss, denoted as $L_{bm}^C$, and path loss, the sum data transmission rate to all targets within the current time frame is quantified as follows:
\begin{equation}
    R(\{\tau_t^n\}_{n=1}^N) = \tau_t^c B\sum_{n=1}^N \log_2\left(1+\frac{P_t L L_{bm}^C}{\sigma^2}\right)
    \label{comm}
\end{equation}
where $B$ represents the bandwidth, $P_t$ is the transmission power, and $L$ is the path loss, defined as:
\begin{equation}
    L = \left(\frac{d_0}{d}\right)^{\eta/2}
\end{equation}
where $d_0$ is a reference distance and $d$ is the actual distance from the radar to the target, with $\eta$ being the path loss factor.

Similar to the definition of beam misalignment loss in \eqref{BM}, the communication beam misalignment loss $L_{bm}^C$ is defined as
\begin{equation}
L_{bm}^C = 
\begin{cases} 
  \cos^i(|\Theta - \hat{\Theta}|) & \text{if } |\Theta - \hat{\Theta}|\leq\frac{\pi}{2} \\
  0 & \text{if } |\Theta - \hat{\Theta}|>\frac{\pi}{2}
\end{cases}.
\label{BM_c}
\end{equation}

The objective is to find a time allocation policy $\pi$ that can efficiently allocate the available time between tracking and communication to maximize the sum rate received by the targets. The radar must allocate sufficient time for accurate target location estimation, thereby improving both $L_{bm}^T$ and $L_{bm}^C$. At the same time, adequate time $\tau_t^c$ should be dedicated to communication since the objective function $R$ is linearly proportional to $\tau_t^c$.

Following \cite{altman1999constrained} and \cite{tessler2018reward}, the problem considered in this paper can be formulated as the following constrained optimization problem:
\begin{equation}
\begin{aligned}
& \underset{\pi}{\text{maximize}}
& & \sum_{m=0}^\infty \gamma^m R(\{\tau_{t+m}^n\}_{n=1}^N) \\
& \text{subject to}
& & \sum_{m=0}^\infty \gamma^m \left(\sum_{n=1}^N\tau_{t+m}^n -T_0\right)\leq 0.\\
\end{aligned}
\label{optimization}
\end{equation}

Utilizing Lagrangian relaxation, the budget constraint specified in (\ref{optimization}) can be integrated into the objective function. This is achieved by introducing a non-negative dual variable, denoted as $\lambda_t$. Consequently, problem (\ref{optimization}) is transformed into an unconstrained optimization problem, as detailed below:

\begin{align}
	\min_{\lambda_t \geq 0} \max_{\pi} \sum_{m=0}^\infty \gamma^m \Bigg[R(\{\tau_{t+m}^n\}_{n=1}^N)-\lambda_t\left(\sum_{n=1}^N \tau_{t+m}^n - T_0\right)\Bigg].
	\label{dual}
\end{align}

The optimization problem expressed in (\ref{dual}) serves as the dual problem to the primary one outlined in (\ref{optimization}). In convex optimization scenarios, the duality gap between these two problems vanishes. It's important to recognize that $\lambda_t$ is dynamic over time, and arbitrarily setting its value might result in a suboptimal solution. The forthcoming section introduces a constrained deep reinforcement learning approach to effectively tackle this optimization challenge.

\section{Constrained Deep Reinforcement Learning}

In this work, we propose a constrained deep reinforcement learning (CDRL) framework to tackle the time allocation in multi-target tracking and communication in an ISAC system. The goal is to maximize the total communication quality between the base station and the targets, while the total time budget is constrained below a certain threshold. We utilize deep Q-learning (DQL) to achieve this goal.

There is a single DQN in the proposed framework and $N$ tracking tasks. At each time slot $t$, each task $n$ needs to decide its dwell time $\tau_t^n$ for tracking target $n$ with the common DQN. After deciding the dwell time $\{\tau_t^n \}_{n=1}^N$, the communication time $\tau_t^c$ is computed, and the sum rate $R$ of communication can be computed according to (\ref{comm}).

\subsection{State}
In this study, the state of task $ n $ at time slot $t$, denoted as $s_t^n$, is defined as follows:
\begin{equation}
    s_t^n = [ \{\sigma^2_{\theta_{t-1}^n}\}_{n=1}^N, \{\tau_{t-1}^n\}_{n=1}^N, \lambda_{t-1} ].
\end{equation}

Here, $ \{\sigma^2_{\theta_{t-1}^n}\}_{n=1}^N $ represents the variance of the estimated azimuth angles of the targets from the preceding time slot. It is important to note that these values are not directly accessible to the agent; instead, they are derived using the Monte Carlo method. This involves generating a multitude of hypothetical target positions ($x, y$) based on the covariance matrix from the extended Kalman Filter, which is part of $ P_{t|t} $ in each time slot. By calculating the azimuth angle $ \theta $ for these positions and analyzing the spread of these values, we obtain the variance $ \sigma^2_{\theta} $. In the training phase, the actual $ x $ and $ y $ values of a target are taken as their mean values, whereas during the testing phase, the estimated $ x $ and $ y $ are used.

The second component, $ \{\tau_{t-1}^n\}_{n=1}^N $, is an array representing the dwell times chosen in the previous time slot. The final term, $ \lambda_{t-1} $, indicates the dual variable from the previous time slot. The total size of the state vector is  $2N + 1 $, encompassing these elements.

\subsection{Action}
Action $a_t^n$ is the dwell time selected for tracking target $n$ in time slot $t$. Each task $n$ can choose a dwell time $\frac{\tau_t^n}{T_0} \in [0, 1]$. We quantize the range into ten levels and hence $a_t^n = \frac{\tau_t^n}{T_0} \in \{0, 0.1, ..., 1\}$. The action is selected by the $\epsilon$-greedy method. The action is either determined by the output of the DQN with probability $(1-\epsilon)$ or randomly sampled from the action space with probability $\epsilon$.

\subsection{Reward}
All the tasks jointly maximize a global reward $r_t$ during time slot $t$. And $r_t$ is defined as

\begin{equation}
    r_t = R(\{\tau_t^n\}_{n=1}^N)-\lambda_t\left(\sum_{n=1}^N \tau_{t}^n - T_0\right).
    \label{reward}
\end{equation}

The first term in $r_t$ denotes the objective function that the CDRL algorithm aims to maximize, and the second term addresses the constraint.

We observed that the reward function, which is based on the instantaneous communication quality \( R(\{\tau_t^n\}_{n=1}^N) \), exhibited instability due to the inherent randomness and noise in the communication process. To address this challenge, we have implemented a critical modification to the reward function. Rather than using the instantaneous beam misalignment angle \( |\Theta - \hat{\Theta}| \) as initially defined in (\ref{BM_c}), we have shifted our approach to incorporate the standard deviation of the azimuth angle \( \sigma_\theta \). Consequently, the communication quality \( R(\{\tau_t^n\}_{n=1}^N) \) is recalculated in accordance with (\ref{comm}).

This modification in the reward function is significant. By relying on the standard deviation of the azimuth angle, the reward metric becomes more robust against the fluctuations and uncertainties inherent in the process. This change enhances the stability of the reward function, making it a more reliable indicator of the effectiveness of the selected actions in terms of communication performance. It enables a more consistent and dependable assessment of the action's impact on communication quality, which is crucial for the success of our framework.

\subsection{Update of the CDRL} 
\begin{algorithm}[!ht]
	\caption{CDRL Algorithm}
	\begin{spacing}{1}
	\begin{algorithmic}[1]
		\STATE{Initialize the DQN parameters $\Phi_0^{\pi}$ with random values.}
		\STATE{Initialize states $\{\mathbf{s_0^n}\}_{n=1}^N$ as zero vectors and $\lambda_t$ as $\lambda_0$.}
		\FOR{time slot $t=0,1,..., T_{max}$}
		\FOR{each agent $n = 1, 2,..., N$}
		\STATE{Select an action $a_t^n$ based on the current state $\mathbf{s_t^n}$ with the DQN $\Phi_t^{\pi,n}$ and $\epsilon$-greedy method.}

		\ENDFOR
		\STATE{Compute reward $r_t$ according to (\ref{reward}).}
		\FOR{each agent $n = 1, 2,..., N$}
		\STATE{Store the experience ($\mathbf{s_t^n}$, $a_t^n$, $r_t$, $\mathbf{s_{t+1}^n}$) to the DQN experience buffer.}
		\STATE{Update $\Phi_t^{\pi}$ to $\Phi_{t+1}^{\pi}$ with experience replay and back-propagation.}
		\ENDFOR
		\STATE{$\lambda_{t+1} = \max(0,\; \lambda_t - \alpha (\sum_{n=1}^N \tau_{t}^n - T_0) )$.}
		\ENDFOR
	\end{algorithmic}
	\end{spacing}
	\label{algorithm1}
\end{algorithm}

The proposed CDRL algorithm is described in Algorithm \ref{algorithm1} above. We simultaneously update the DQN parameters $\Phi_t^{\pi}$ and the dual variable $\lambda_t$.

The objective of the proposed CDRL algorithm is to find a solution to the problem in (\ref{dual}). By setting the reward function as (\ref{reward}), the DQN will learn to maximize the discounted reward $r_t$, i.e.

\begin{equation}
    \max_{\pi} \sum_{m=0}^\infty \gamma^m \Bigg[R(\{\tau_{t+m}^n\}_{n=1}^N)-\lambda_t\left(\sum_{n=1}^N \tau_{t+m}^n - T_0\right)\Bigg].
    \label{DQNjob}
\end{equation}

We denote the objective function in (\ref{DQNjob}) as $\mathcal{L}$. Then, the dual variable $\lambda_t$ is updated by minimizing $\mathcal{L}$ over $\lambda_t$, i.e.

\begin{equation}
    \begin{aligned}
	\lambda_{t+1} = \max(0,\; \lambda_t - \alpha \bigtriangledown_{\lambda_t} \mathcal{L}) \\
	= \max\left(0,\; \lambda_t + \alpha \sum_{m=0}^\infty \gamma^m \left(\sum_{n=1}^N \tau_{t+m}^n -T_0\right)\right)
	\end{aligned}
\end{equation}
where $\alpha$ is the learning rate of the dual variable. The gradient $\bigtriangledown_{\lambda_t} \mathcal{L}$ can be estimated with an additional neural network but we simplify use instead
\begin{equation}
	\lambda_{t+1} = \max\left(0,\; \lambda_t + \alpha \left(\sum_{n=1}^N \tau_{t}^n - T_0\right) \right).
\end{equation}

\subsection{Convergence Analysis}
The proposed CDRL algorithm iteratively determines the DQN parameters and the dual variable ($\Phi_t$, $\lambda_t$). Based on the theoretical foundations presented in \cite{tessler2018reward} and Chapter 6 of \cite{borkar2009stochastic}, this iterative approach is identified as a multi-timescale stochastic approximation process, ultimately leading to convergence at a stable point ( $\Phi_t^*, \lambda_t^* $).

\section{Numerical Results}

\begin{table}[!ht]
	\renewcommand{\arraystretch}{1.3}
	
	\caption{Simulation Parameters}
	\label{table1}
	\centering
	\small
	\begin{tabular}{|c||c|}
		\hline $\sigma_{r,0}^2$ ($m^2$)&10\\
		\hline $\sigma_{\theta,0}^2$ $(\text{rad}^2)$&1e-4\\
		\hline $\sigma_w$ ($(m/s^2)^2$) & 5\\
		\hline Reference distance $r_0$ ($m$) & 800\\
		\hline Reference dwell time $\tau_0$ (s) & 2\\
		\hline Revisit interval $T_0$ (s) & 3\\
        \hline Transmit Power $P_t$ (W) & 1\\
        \hline Noise Level $\sigma$ & 0.1\\
        \hline Reference Distance in Path Loss $d_0$ (m) & 500\\
        \hline Bandwidth $B$ (Hz) & 500\\
		\hline DRL discount factor $\gamma$ & 0.9\\
		\hline DRL mini-batch size & 32\\
            \hline Replay memory buffer size & 50000\\
		\hline Exploring probability $\epsilon$ & 0.1\\
		\hline Initial dual variable ($\lambda_0$) & 100\\
		\hline Step size of dual variable ($\alpha$) & 10\\    
		\hline
		
	\end{tabular}
	\label{table1}
\end{table}

\subsection{Experimental Setup}
This section outlines the experimental framework, including the specific hyperparameters utilized for both the environment and the CDRL algorithm. These parameters are detailed in Table \ref{table1}. The architecture of the DQN employed in our study is designed with two hidden layers, each consisting of 64 neurons. To facilitate effective neural transmission and nonlinear modeling between layers, the ReLU (Rectified Linear Unit) activation function is employed.

Our experimental model presupposes a scenario where a maximum of four targets are present. During the training phase of the CDRL algorithm, these targets are introduced into the system at locations and time slots that are generated randomly, thus simulating a dynamic and unpredictable environment that closely mimics real-world conditions.

\subsection{Testing Results}

\begin{figure}[h!]
    \centering
    \includegraphics[width=.5\textwidth]{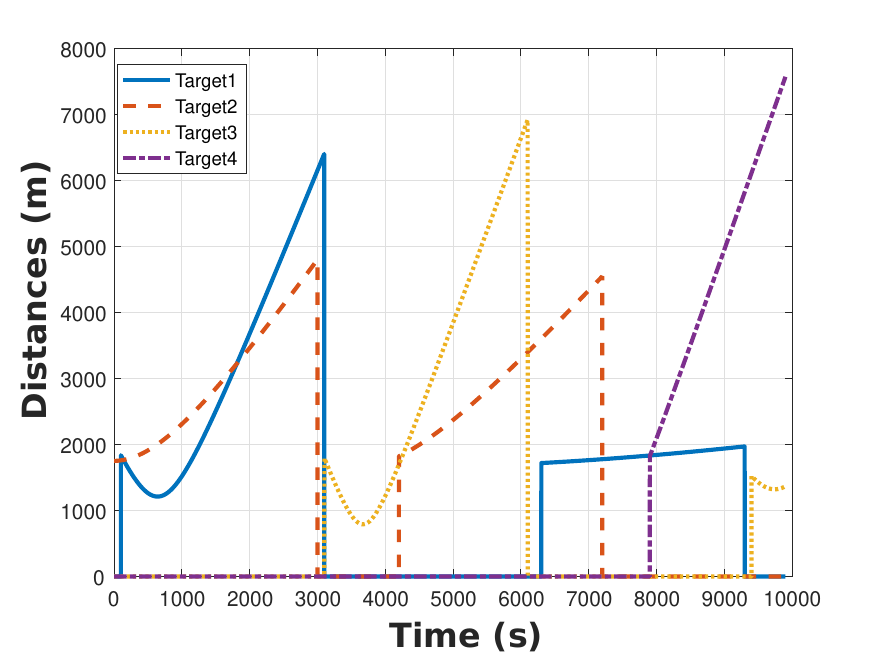}
    \caption{Target distances to the radar}
    \label{distance}
\end{figure}

\begin{figure}[h!]
    \centering
    \includegraphics[width=.5\textwidth]{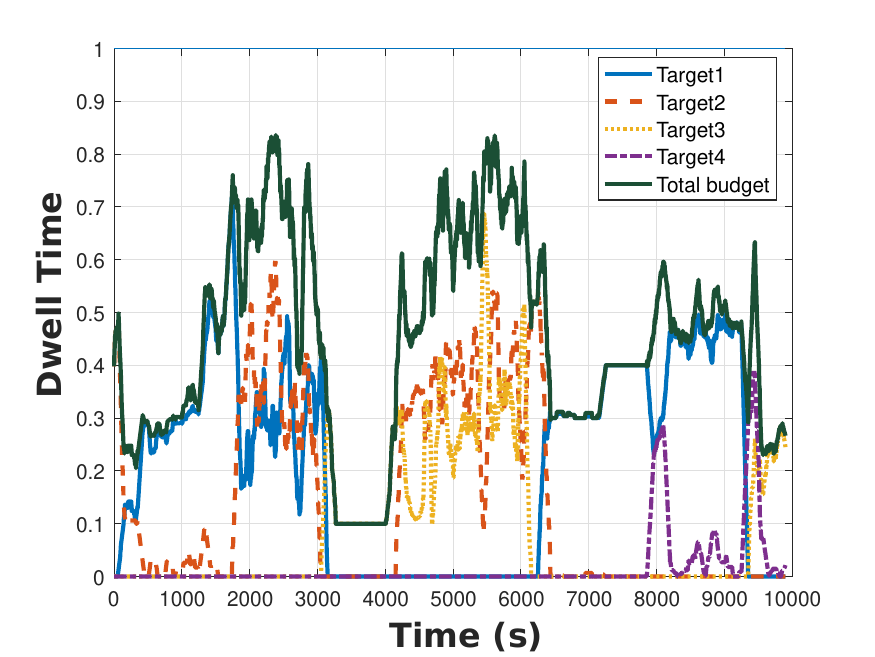}
    \caption{Dwell time allocation strategy learned by CDRL}
    \label{time}
\end{figure}

\begin{figure}[h!]
    \centering
    \includegraphics[width=.5\textwidth]{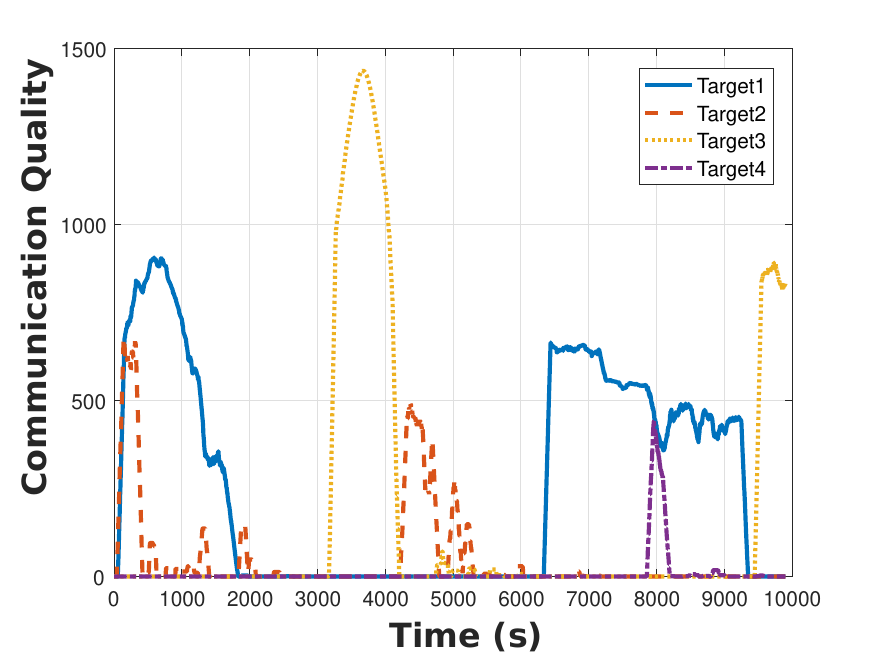}
    \caption{Communication quality achieved by CDRL}
    \label{com_perf}
\end{figure}

The target mobility patterns in the testing phase are illustrated in Fig. \ref{distance}, which specifically showcases the varying distances of the targets in one of our designed test scenarios. To enhance the complexity and diversity of the testing environment, a specific protocol is followed where any target remaining in the environment for an excess of three thousand time slots is systematically removed. This approach ensures a constantly evolving test environment, challenging the algorithm to adapt to frequently changing conditions and thereby providing a robust evaluation of its performance.

Fig. \ref{time} presents the dwell time allocation strategy as learned by our proposed learning framework. A noteworthy observation from this figure is the correlation between target distance and time allocation for tracking. Specifically, the framework tends to allocate a higher proportion of time for tracking when all targets are positioned at a greater distance from the radar. This is particularly evident during the intervals [2000, 3000] and [6000, 6500]. The rationale behind this strategy is intuitive: when tracking accuracy becomes critical due to the increased target distances, it is imperative to prioritize tracking over communication to mitigate the risk of losing communication quality due to tracking errors.

Conversely, as depicted in Fig. \ref{time}, during the intervals [1000, 1200], [3700, 4300], and [6700, 7000], we observe a shift in strategy when a target is close to the radar. In such scenarios, the algorithm adapts by allocating more time to the communication phase. This strategic shift is based on the understanding that enhanced communication time directly and linearly improves the sum rate of data transmission. In these instances, further time investment in tracking yields diminishing returns in terms of communication quality improvement. This adaptive time allocation demonstrates the framework's capacity to dynamically balance the time allocated for tracking and communication based on the real-time dynamic environment. Figure \ref{com_perf} illustrates the individual communication rate (to each target) achieved by the CDRL during this test case.

The comparative analysis of the sum-rate performance between our proposed algorithm and various benchmark strategies is detailed in Table \ref{sumrate}. As an example, the strategy labeled `Fixed 0.1' refers to a fixed dwell time allocation protocol. In this strategy, whenever a target is present in the environment, a consistent dwell time of \(0.1T_0\) is allocated for tracking it. Subsequently, after assigning the specified tracking time to each detected target, the remaining time within the time slot is dedicated exclusively to the communication phase. This comparison reveals that the proposed CDRL algorithm consistently outperforms the pre-determined fixed allocation schemes in terms of communication efficiency. Notably, while the `Fixed 0.2' strategy exhibits performance metrics closely rivaling that of CDRL in this specific tested scenario, it is important to recognize the inherent limitations of such fixed schemes. Specifically, the `Fixed 0.2' strategy lacks the adaptive capabilities essential for optimizing performance in dynamic and unpredictable environments, a critical advantage that our CDRL algorithm inherently possesses.

\begin{table}[h]
\centering
\begin{tabular}{|c|c|c|}
\hline
 & \textbf{Avg. Sum-rate} & \textbf{Percentage} \\ \hline
CDRL           &   466.37      &    100\%   \\ \hline
Fixed 0.1  &    345.94      &    74.18\%  \\ \hline
Fixed 0.2  &      449.42    &     96.37\% \\ \hline
Fixed 0.3  &       383.02   &    82.13\% \\ \hline
\end{tabular}
\vspace{.3cm}
\caption{Sum-rate Comparison}
\label{sumrate}
\end{table}

\section{Conclusions}

This study introduces a novel CDRL framework tailored for efficient time allocation in ISAC systems. The CDRL framework innovatively integrates the simultaneous updating of neural network parameters and the dual variable in order to find a strategic balance to optimize communication performance within predefined budget constraints. Our numerical results are demonstrative of the CDRL algorithm's capability to intelligently learn and implement a balanced time allocation strategy. This strategy adeptly navigates the trade-off between tracking accuracy and communication efficiency. A key highlight of our findings is the superior sum rate performance of the proposed CDRL framework when compared to several arbitrarily-designed fixed allocation strategies. This enhanced performance underscores the efficiency of the CDRL in adapting to dynamic environmental conditions and aligning resource allocation in favor of maximized communication throughput.

\bibliographystyle{IEEEtran}
\bibliography{ref}

\begin{thebibliography}{10}
\providecommand{\url}[1]{#1}
\csname url@samestyle\endcsname
\providecommand{\newblock}{\relax}
\providecommand{\bibinfo}[2]{#2}
\providecommand{\BIBentrySTDinterwordspacing}{\spaceskip=0pt\relax}
\providecommand{\BIBentryALTinterwordstretchfactor}{4}
\providecommand{\BIBentryALTinterwordspacing}{\spaceskip=\fontdimen2\font plus
\BIBentryALTinterwordstretchfactor\fontdimen3\font minus
  \fontdimen4\font\relax}
\providecommand{\BIBforeignlanguage}[2]{{%
\expandafter\ifx\csname l@#1\endcsname\relax
\typeout{** WARNING: IEEEtran.bst: No hyphenation pattern has been}%
\typeout{** loaded for the language `#1'. Using the pattern for}%
\typeout{** the default language instead.}%
\else
\language=\csname l@#1\endcsname
\fi
#2}}
\providecommand{\BIBdecl}{\relax}
\BIBdecl

\bibitem{orman1996scheduling}
A.~Orman, C.~N. Potts, A.~Shahani, and A.~Moore, ``Scheduling for a
  multifunction phased array radar system,'' \emph{European Journal of
  Operational Research}, vol.~90, no.~1, pp. 13--25, 1996.

\bibitem{butler1997resource}
J.~Butler, A.~Moore, and H.~Griffiths, ``Resource management for a rotating
  multi-function radar,'' in \emph{Radar 97 (Conf. Publ. No. 449)}.\hskip 1em
  plus 0.5em minus 0.4em\relax IET, 1997, pp. 568--572.

\bibitem{deligiannis2017game}
A.~Deligiannis, A.~Panoui, S.~Lambotharan, and J.~A. Chambers, ``Game-theoretic
  power allocation and the nash equilibrium analysis for a multistatic {MIMO}
  radar network,'' \emph{IEEE Transactions on Signal Processing}, vol.~65,
  no.~24, pp. 6397--6408, 2017.

\bibitem{charlish2020development}
A.~Charlish, F.~Hoffmann, C.~Degen, and I.~Schlangen, ``The development from
  adaptive to cognitive radar resource management,'' \emph{IEEE Aerospace and
  Electronic Systems Magazine}, vol.~35, no.~6, pp. 8--19, 2020.

\bibitem{haykin2006cognitive}
S.~Haykin, ``Cognitive radar: a way of the future,'' \emph{IEEE Signal
  Processing Magazine}, vol.~23, no.~1, pp. 30--40, 2006.

\bibitem{liu2022survey}
A.~Liu, Z.~Huang, M.~Li, Y.~Wan, W.~Li, T.~X. Han, C.~Liu, R.~Du, D.~K.~P. Tan,
  J.~Lu \emph{et~al.}, ``A survey on fundamental limits of integrated sensing
  and communication,'' \emph{IEEE Communications Surveys \& Tutorials},
  vol.~24, no.~2, pp. 994--1034, 2022.

\bibitem{liu2022integrated}
F.~Liu, Y.~Cui, C.~Masouros, J.~Xu, T.~X. Han, Y.~C. Eldar, and S.~Buzzi,
  ``Integrated sensing and communications: Toward dual-functional wireless
  networks for {6G} and beyond,'' \emph{IEEE Journal on Selected Areas in
  Communications}, vol.~40, no.~6, pp. 1728--1767, 2022.

\bibitem{schope2021constrained}
M.~Sch{\"o}pe, H.~Driessen, and A.~Yarovoy, ``A constrained {POMDP} formulation
  and algorithmic solution for radar resource management in multi-target
  tracking,'' \emph{ISIF Journal of Advances in Information Fusion}, vol.~16,
  no.~1, p.~31, 2021.

\bibitem{de2021radar}
T.~de~Boer, M.~I. Sch{\"o}pe, and H.~Driessen, ``Radar resource management for
  multi-target tracking using model predictive control,'' in \emph{2021 IEEE
  24th International Conference on Information Fusion (FUSION)}.\hskip 1em plus
  0.5em minus 0.4em\relax IEEE, 2021, pp. 1--8.

\bibitem{mnih2015human}
V.~Mnih, K.~Kavukcuoglu, D.~Silver, A.~A. Rusu, J.~Veness, M.~G. Bellemare,
  A.~Graves, M.~Riedmiller, A.~K. Fidjeland, G.~Ostrovski \emph{et~al.},
  ``Human-level control through deep reinforcement learning,'' \emph{Nature},
  vol. 518, no. 7540, pp. 529--533, 2015.

\bibitem{thornton2020deep}
C.~E. Thornton, M.~A. Kozy, R.~M. Buehrer, A.~F. Martone, and K.~D. Sherbondy,
  ``Deep reinforcement learning control for radar detection and tracking in
  congested spectral environments,'' \emph{IEEE Transactions on Cognitive
  Communications and Networking}, vol.~6, no.~4, pp. 1335--1349, 2020.

\bibitem{selvi2020reinforcement}
E.~Selvi, R.~M. Buehrer, A.~Martone, and K.~Sherbondy, ``Reinforcement learning
  for adaptable bandwidth tracking radars,'' \emph{IEEE Transactions on
  Aerospace Electronic Systems}, vol.~56, no.~5, pp. 3904--3921, 2020.

\bibitem{meng2021deep}
F.~Meng, K.~Tian, and C.~Wu, ``Deep reinforcement learning-based radar network
  target assignment,'' \emph{IEEE Sensors Journal}, vol.~21, no.~14, pp.
  16\,315--16\,327, 2021.

\bibitem{koch1999adaptive}
W.~Koch, ``Adaptive parameter control for phased-array tracking,'' in
  \emph{Signal and Data Processing of Small Targets 1999}, vol. 3809.\hskip 1em
  plus 0.5em minus 0.4em\relax SPIE, 1999, pp. 444--455.

\bibitem{meikle2008modern}
H.~Meikle, \emph{Modern radar systems}.\hskip 1em plus 0.5em minus 0.4em\relax
  Artech House, 2008.

\bibitem{welch1995introduction}
G.~Welch, G.~Bishop \emph{et~al.}, ``{An introduction to the Kalman filter},''
  1995.

\bibitem{altman1999constrained}
E.~Altman, \emph{Constrained Markov decision processes}.\hskip 1em plus 0.5em
  minus 0.4em\relax CRC Press, 1999, vol.~7.

\bibitem{tessler2018reward}
C.~Tessler, D.~J. Mankowitz, and S.~Mannor, ``Reward constrained policy
  optimization,'' \emph{arXiv preprint arXiv:1805.11074}, 2018.

\bibitem{borkar2009stochastic}
V.~S. Borkar, \emph{Stochastic approximation: a dynamical systems
  viewpoint}.\hskip 1em plus 0.5em minus 0.4em\relax Springer, 2009, vol.~48.

\end{thebibliography}
\end{document}